\documentclass[final]{cvpr}

\usepackage{times}
\usepackage{epsfig}
\usepackage{graphicx}
\usepackage{amsmath}
\usepackage{amssymb}

\usepackage{subfigure}
\usepackage{bbm}
\usepackage{amsfonts}
\usepackage{enumerate}
\usepackage{tabu}
\usepackage{multirow}
\usepackage{makecell}
\usepackage[ruled]{algorithm2e}
\usepackage{color}
\newcommand{\tabincell}[2]{\begin{tabular}{@{}#1@{}}#2\end{tabular}}
\newcommand{\red}[1]{\textcolor{red}{#1}}
\newcommand{\green}[1]{\textcolor{green}{#1}}
\newcommand{\blue}[1]{\textcolor{blue}{#1}}
\usepackage[pagebackref=true,breaklinks=true,colorlinks,bookmarks=false]{hyperref}

\def\cvprPaperID{6052} 
\def\confYear{ICCV 2021}

\begin{document}

\title{Real-Time Visual Object Tracking via Few-Shot Learning}

\setcounter{secnumdepth}{2}
\author{Jinghao Zhou\textsuperscript{\rm 1,2}, Bo Li\textsuperscript{\rm 2}, Peng Wang\textsuperscript{\rm 1}, Peixia Li\textsuperscript{\rm 3}\\
Weihao Gan\textsuperscript{\rm 2}, 
Wei Wu\textsuperscript{\rm 2}, Junjie Yan\textsuperscript{\rm 2},
Wanli Ouyang\textsuperscript{\rm 3}\\
\textsuperscript{\rm 1}Northwestern Polytechnical University\\
\textsuperscript{\rm 2}SenseTime Group Limited\\
\textsuperscript{\rm 3}University of Sydney\\
\tt\small jensen.zhoujh@gmail.com, \{libo,ganweihao\}@sensetime.com,\\
\tt\small peng.wang@nwpu.edu.cn, \{peixia.li,wanli.ouyang\}@sydney.edu.au
}

\maketitle

\begin{abstract}
Visual Object Tracking (VOT) can be seen as an extended task of Few-Shot Learning (FSL). While the concept of FSL is not new in tracking and has been previously applied by prior works, most of them are tailored to fit specific types of FSL algorithms and may sacrifice running speed.
In this work, we propose a generalized two-stage framework that is capable of employing a large variety of FSL algorithms while presenting faster adaptation speed.
The first stage uses a Siamese Regional Proposal Network to efficiently propose the potential candidates and the second stage reformulates the task of classifying these candidates to a few-shot classification problem. Following such a coarse-to-fine pipeline, the first stage proposes informative sparse samples for the second stage, where a large variety of FSL algorithms can be conducted more conveniently and efficiently. 
As substantiation of the second stage, we systematically investigate several forms of optimization-based few-shot learners from previous works with different objective functions, optimization methods, or solution space. Beyond that, our framework also entails a direct application of the majority of other FSL algorithms to visual tracking, enabling mutual communication between researchers on these two topics. 
Extensive experiments on the major benchmarks, VOT2018, OTB2015, NFS, UAV123, TrackingNet, and GOT-10k are conducted, demonstrating in desirable performance gain and a real-time speed. 
\end{abstract}

\section{Introduction}


The fast adaptivity of a tracker lies crucial in handling sequential variation of the target, especially in terms of the discriminability against noisy distraction. The task setting of tracking, using very few data for classification in a limited time budget, closely resembles the task setting of few-shot learning (FSL). With the prosperous development and superior performance of a large variety of newly-proposed FSL algorithms, an idea of visual tracking via few-shot learning arises naturally.
In few-shot learning, we hypothesize an existence of meta-knowledge shared across training tasks and new tasks, which, in the field of visual tracking, can be interpreted as the model's adaptivity to track any unseen object of interest across all sequences.

The concept of FSL, however, is not new and has been introduced by several previous works  \cite{MetaTracker,UnifiedDet,Det-MAML,ROAM,DiMP}.
By involving the online updating into the offline training stage as an inner loop, these methods turn the manual design of the online update strategy \cite{KCF, MDNet} into a data-driven module. 
Compared to siamese networks \cite{SiamFC, SiamRPN}, which can be regarded as metric-based few-shot learners via pure matching, the existing applications of FSL algorithms in tracking mainly focus on optimization-based few-shot learners with its powerful adaptation capability in distinguishing novel classes.
However, previous methods hinge greatly on specific FSL algorithms, the majority of which resorts to a similar design as MAML \cite{MAML} by learning a gradient-descent strategy with trainable parameters. Specifically, most of these approaches limit themselves to an optimization design of specific convolution kernels over the whole image instead of more customized weight learning (e.g, such as a matrix multiplication factor) with sparse samples as in FSL's task setting, which prohibits a direct introduction of various new FSL algorithms, since a direct application of various FSL algorithms by taking all locations of the whole image as input samples will surely sacrifice its tracking speed.

Based on the previous analysis, the cardinal motivation of this paper is to design a universal framework for real-time tracking via few-shot learning, where more customized learning can be conducted on sparse samples. A cascaded framework \cite{fasterrcnn} thus meets our requirements by first proposing several potential target regions as sparse candidates to allow a few-shot learning over them.
Similar to previous studies, this work mainly focuses on the discussion of optimization-based few-shot learning methods. While any other type of few-shot learners can suit well in our framework, the optimization-based approaches excel at its strong discriminability among novel classes in low data regime thus entail better class boundaries in real-time tracking. 
Following the state-of-the-art approaches in FSL, the optimization-based few-shot learners can be implemented as either as a stepwise gradient descender (GD) \cite{MAML,Reptile} or a differentiable quadratic programming (QP) solver \cite{MetaOptNet,R2D2}, which can be further inserted into network as a layer enabling an end-to-end training with feature extractor.

To this end, we design a generalized two-stage cascaded framework, where the first stage efficiently proposes potential target regions while the second stage is online optimized yielding target-specific weights for few-shot learning.
The proposed coarse-to-fine cascaded pipeline lies crucial in efficient and effective tracking since the first stage filters out uninformative easy negatives thus few-shot learning is only performed on sparse informative samples. 
In practice, while the first stage is limited to any specific method, we directly take recently dominant siamese trackers \cite{SiamRPN++} as the first stage considering its high computational efficiency and great proposal ability.
Further, we instantiate a number of few-shot learners in the second stage with various formulations (e.g, objective, space, and optimization methods etc.), validating and analyzing their few-shot learning ability and computation efficiency. 
Our proposed method can operate at a $40 \sim 60$ Frame-Per-Second (FPS) beyond real-time requirements.
Empirical results on the major benchmarks: VOT2018 \cite{vot2018}, OTB2015 \cite{otb2015}, NFS \cite{nfs}, UAV123 \cite{uav123}, TrackingNet \cite{trackingnet}, and GOT-10k \cite{got10k} verify the effectiveness of our proposed method.

\section{Related Work}

\noindent\textbf{Visual Tracking.} Visual object tracking is traditionally categorized into two groups in general. Generative trackers \cite{SINT,SiamFC} base on the matching results of the features following a non-parametric nearest-neighbor methodology, while discriminative trackers with either tracking-by-detection framework \cite{FCNT,MDNet} or correlation filter \cite{KCF,ECO} resort to an online updated parametric classifier. A related study \cite{DROL} shows that generative trackers prevail given its generative embedding space crucial for high-fidelity representation \cite{SiamRPN,SiamMask}, whilst discriminative trackers \cite{MOSSE,ATOM,DiMP} exploit the background information in context to learn a discriminant model thus perform well at suppressing the distractors. 

\noindent\textbf{Cascaded Framework for Tracking.} Concurrent to our work, several recent work \cite{CRPN,SPM} has investigated the cascaded framework into the siamese network. While either introducing an extra relation network \cite{SPM} to distinguish target and background or stacking several regional proposal networks \cite{CRPN} for gradual discrimination on the target, the innate mechanism of pure matching and the absence of online update heavily limits the tracker's performance in dealing with distractors. 
Comparing to their works, our second stage takes the samples proposed by the first stage and solve a few-shot classification problem with optimization-based few-shot learners. With the maintenance of an online support set during tracking to store target deformation and recurrent optimization of a target-specific weight, our method greatly compensates for siamese discriminative capability.

\noindent\textbf{Few-Shot Learning.} As another significant matter in this work, few-shot learning has been thoroughly studied in the field of few-shot classification. While the early study focuses on generative models \cite{fei2006one} to infer the data distribution from limited resources, recent years have seen surging researches based on discriminative models, which popularly resort to a meta-learning framework by its core. Through learning to learn, the transferable knowledge can be extracted from the distribution of auxiliary sub-tasks. Among all the proposed work, optimization-based approaches \cite{GDRNN,MAML} learn to update the model's weights on a sparse set of data by gradient descent such that meta-information remains in the form of parameter initialization or updating strategy. Comparatively speaking, metric-based methods \cite{siamese,prototypical,matching} integrate the procedure of training and testing into a projection function such that when represented in this embedding space, where similar schematics are adjacent. Model-based methods \cite{SNAIL,MANN,MetaNet} leans a predict model parameters through a parameterized predictors.

\noindent\textbf{Few-Shot Learning for Tracking.} Given the recent prosperity of few-shot learning, several recent works formulate visual tracking into a few-shot learning problem. We note that the popular siamese framework \cite{SiamFC, SiamRPN, SiamRPN++} by its nature is a metric-based model by learning a distance-based prediction rule within the embedding space. Though with simple classification rules, these non-parametric models exclude a mechanism for feature selection thus are not robust to noisy features. 
On the contrary, optimization-based model \cite{DiMP,MetaTracker,Det-MAML,UnifiedDet} uses an explicit gradient-descent algorithm to adjust the parameters of the model given an online sampled dataset. Model-based model \cite{ROAM,GradNet,CLNet} learns a parameterized predictor to estimate model parameters by implicitly leveraging gradient or latent distribution as meta information. The majority of these methods utilize the gradient under certain objectives for the online model update. Some approaches \cite{ROAM,GradNet,DiMP} are optimized under L2 or hinge loss thus form a ridge regression problem where the decision boundary is linear. However, ridge regression is not robust to outliers and unable to pick informative or decisive hard examples. In comparison, several other approaches \cite{Det-MAML, UnifiedDet} are optimized under cross-entropy loss thus form a logistic regression problem. Though more complicated the non-linear model, these methods often agonize over slow convergence and over-fitting.

\section{Method}

\subsection{Cascaded Tracking}

\begin{figure*}[t]
\centering
\includegraphics[width=16cm]{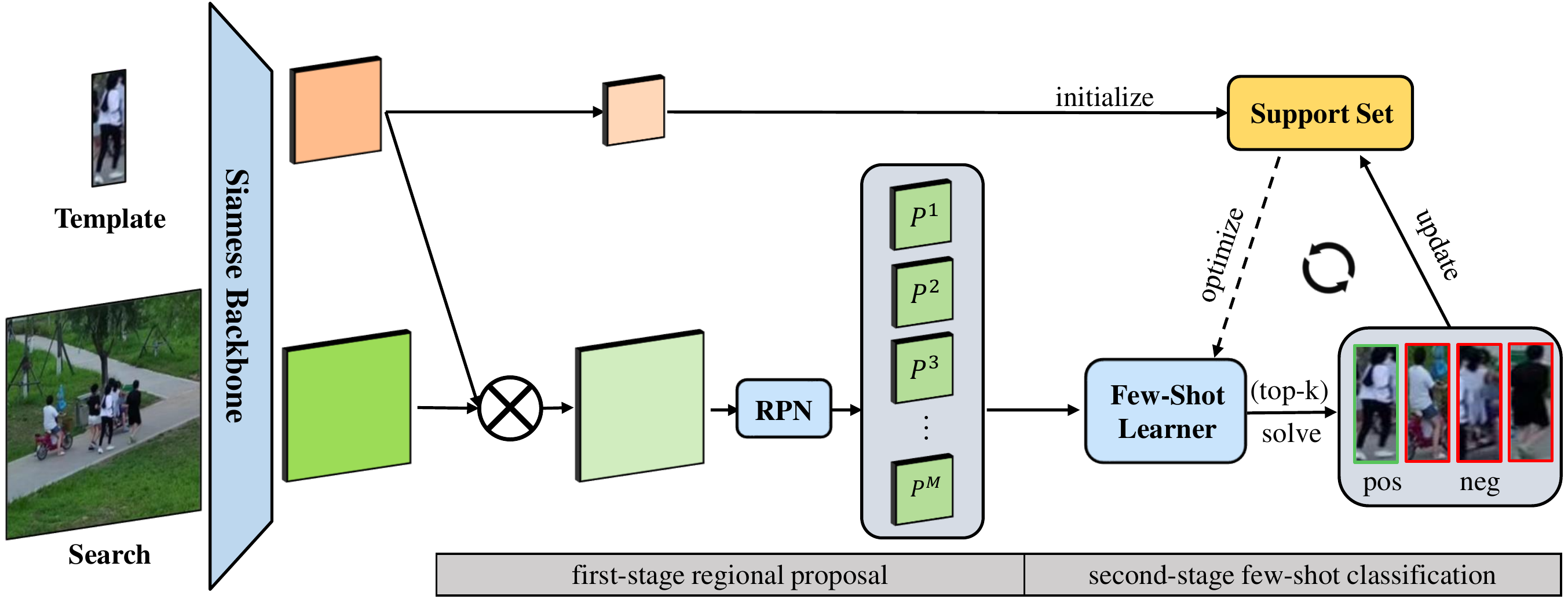}
\caption{Overview of the network architecture (only the classification branch for second stage is showcased for brevity). The first stage proposes total $M$ candidates per frame upon which few-shot classification are conducted. The top-$4$ candidates with highest classification scores are stored into support set to optimize the few-shot learner in subsequent frames.}
\label{architecture}
\vspace{-0.5cm}
\end{figure*}

The proposed framework is a cascaded two-stage architecture. Specifically, the first stage is a Region Proposal Network (RPN) which first proposes potential candidates with similar semantics to the template. To guarantee a real-time speed and high recall rate, we choose SiamRPN++ \cite{SiamRPN++} considering its high efficiency and great proposal ability. We recommend readers refer to \cite{SiamRPN,SiamRPN++} for more details. 
All the proposals are then extracted by RoI Align \cite{maskrcnn} and further forwarded to the second stage, where the one with the highest confidence is selected as the target in the current frame while the others as distractors. A visual overview of our overall architecture is provided in Figure \ref{architecture}. 

We formulate the second-stage as a $N$-shot-$2$-way few-shot classification problem, where $N$ is the number of samples in the support set we maintain during tracking. We note that the chosen first-stage SiamRPN++ and a large amount of its variants \cite{SPM, SiamMask} can be considered as metric-based few-shot learners, where the classification score is derived through similarity matching.
As substantiation of the second stage, we look into several optimization-based few-shot learners for classification, as they often lend well to out-of-distribution tasks and handle well to sequentially varying $N$. All the stored candidates are labeled with an online-generated pseudo label and optimized under certain objectives to obtain a sequence-specific classification weight. 

The proposed coarse-to-fine framework enables a real-time few-shot learning for visual tracking, where online learning is conducted only over informative sparse instances rather than the whole search regions. For the latter, a large number of easy negatives may dominate the learning process, acquiring a sub-par discriminative model and inflicting a computation burden. 
Next, in Section \ref{sec:secondstage}, we shall first introduce how few-shot learning is conducted in our framework. In Section \ref{sec:optimization}, we then investigate and compare several optimization-based few-shot learners, and finally introduce the complete tracking pipeline in Section \ref{sec:tracking}. 

\subsection{Second Stage as Few-Shot Learning}
\label{sec:secondstage}

Given the training set $Z=\{(x_n,y_n)\}_{n=1}^T \sim\delta$ and testing set $Z'=\{(x'_n,y'_n)\}_{n=1}^Q \sim \delta$ sampled independently from the distribution $\delta$, the few-shot learner $\Lambda$ is designed to estimate parameters $\theta$ on training set $\phi(Z;\varphi)$, where $\phi(Z;\varphi)\triangleq \{(\phi(x_n;\varphi),y_n)\}$, so that it generalized well to the unseen testing set $\phi(Z';\varphi)$. The $y_n$ is one-hot encoding for target class. Formally, the objective for minimizing generalization error $L_{meta}$ can be formulated as
\begin{equation}
\label{eq:metaloss}
    \min_{\varphi,\rho}\mathbb{E}_{Z,Z'}[L_{meta}(\phi(Z'; \varphi);\theta), \ \mathrm{where} \ \theta = \Lambda(\phi(Z; \varphi); \rho)],
\end{equation}
where $\theta$ is the sequence-specific parameter and is obtained in the meta-training stage (inner loop) under the base loss $L_{base}$, while $\varphi$ is the parameter of the feature embedding and $\rho$ corresponds to the hyper-parameters of the few-shot learner $\Lambda$, which are updated via stochastic gradient descent (SGD) in the meta-testing (outer loop) stage. 

To facilitate the learning of both feature embedding and adaptation, we resort to a recently proposed Focal Loss \cite{RetinaNet} for outer loop to alleviate the problem of class imbalance, $L_{meta}$ can be expressed as
\begin{equation}
\begin{aligned}
\label{eq:focalloss}
    L_{meta}(\phi(Z')) = -\sum_{(x',y')} &\alpha[1-\mathrm{softmax}(\gamma \cdot \theta^T\phi(x'))]^\beta \cdot \\
    & y' \cdot \mathrm{log}[\mathrm{softmax}(\gamma \cdot \theta^T\phi(x'))]
\end{aligned}    
\end{equation}
where $\gamma$ is a learnable scaling factor and $\theta^T$ are weight for the background and foreground class. $\alpha$ and $\beta$ are hyper parameters of focal loss. $\varphi$ and $\theta$ are discarded for the simplicity of notation. 

The choice of the few-shot learner $\Lambda$ has a significant impact on Equation \ref{eq:metaloss}. The few-shot learner that computes $\theta = \Lambda(\phi(Z; \varphi); \rho)$ has to be discriminative enough to distinguish the instances proposed in the first stage. Specifically, optimization-based few-shot learners are mainly considered given its strong discriminability and clear adaptation dynamics. In the next section, we shall discuss the property (e.g, classification ability, computation cost, optimization methods, etc.) of several few-shot learners and how the optimization is conducted.

\subsection{Optimization-Based Few-Shot Learners}
\label{sec:optimization}

Next, we showcase various choices as the substantiations of the optimization-based few-shot learner for the second stage. In this work, we mainly consider $\Lambda$ as linear convex classifiers, where the few-shot learner's objective is convex for effective online learning. As demonstrated later, other genres of few-shot learners, such as metric-based few-shot learning algorithms also suits well in our framework.

\noindent\textbf{RR-prim-itr} \cite{MAML,DiMP,ROAM} can be seen as a special case of MAML \cite{MAML}, which solves ridge regression (RR) in primal space iteratively with $L_{base}$ specifying as $L2$ loss. While such practice has been popularly applied in the field of visual tracking \cite{DiMP,ROAM}, the problem in our framework can be expressed as 
\begin{equation}
\begin{aligned}
\label{eq:rr-prim-itr}
    \Lambda(\phi(Z); \theta) &= \mathop{\arg\min}_{\theta} \sum_{n=1}^N ||\omega_n (\theta^T\phi(x_n)-y_n)||^2 + \lambda ||\theta||^2 \\
    &= \mathop{\arg\min}_{\theta} ||\Phi \theta-\boldsymbol{y}||^2 + \lambda ||\theta||^2
\end{aligned}   
\end{equation}
where $\omega_n$ is the weighting term for each sample. The $n^{th}$ row of $\Phi$ is given by $\omega_n \phi(x_n)$ and $\boldsymbol{y}=(\omega_1 y_1, ...,\omega_N y_N)^T$ with $y_n$ being an one-hot vector of the given label. $\lambda$ is the regularization term. To accelerate convergence speed, we solve the above objective with steepest descent methodology \cite{DiMP}, which scales the step length with its Hessian matrix $Q$. The problem of RR results in a quadratic program (QP), such that $Q$ degenerates to the inner product of Jacobian vector as $Q=J^TJ$. The few-shot learner thus update as follows:
\begin{equation}
\label{eq:rr-prim-itr-update}
    \theta^{(i+1)} = \theta^{(i)} - \frac{\triangledown L_{meta}(\theta^{(i)})^T\triangledown L_{meta}(\theta^{(i)})}{\triangledown L_{meta}(\theta^{(i)})^TQ\triangledown L_{meta}(\theta^{(i)})} \triangledown L_{meta}(\theta^{(i)})
\end{equation}
where $i$ denotes the number of iterations of stepwise optimization. We initialize the parameter $\theta^{(0)}$ with template feature adaptively pooed to the kernel size.

\noindent\textbf{RR-dual-itr} \cite{MetaOptNet} solves RR in dual space iteratively where $\theta$ are formulated as linear combinations of feature vectors from training set, with dual variable $\boldsymbol{a}$ as weighting factor, which can be formulated as
\begin{equation}
\label{eq:rr-dual-itr-weight}
    \theta(\boldsymbol{a}) = \sum_{n=1}^N a_n\phi(x_n) = \Phi^T \boldsymbol{a}
\end{equation}
where  $\boldsymbol{a} = (a_1, ..., a_N)^T$. By substituting the above equation into Equation \ref{eq:rr-prim-itr}, we have the few-shot learner formulated as
\begin{equation}
\begin{aligned}
\label{eq:rr-dual-itr}
    \Lambda(\phi(Z); \boldsymbol{a}) = &\mathop{\arg\min}_{\boldsymbol{a}} ||\theta(\boldsymbol{a})||^2 + \lambda||\boldsymbol{a}||^2 - 2\boldsymbol{a}^T\boldsymbol{y} \\
    & \mathop{\arg\min}_{\boldsymbol{a}} \boldsymbol{a}^T(\Phi\Phi^T + \lambda I)\boldsymbol{a} - 2\boldsymbol{a}^T\boldsymbol{y}
\end{aligned}
\end{equation}
where the QP is conducted over the dual variables $\boldsymbol{a}\in \mathbb{R}^{N\times2}$ rather than feature vector $\theta\in \mathbb{R}^{d\times2}$. In the low-data regime, considering sample amount $N$ is often much smaller then feature dimension $d$, we solve the QP using a differeniable GPU-based QP solver \cite{OptNet} instead of unrolling the hand-crafted optimization as in Equation \ref{eq:rr-prim-itr}. Given that the complexity of the complete pass for solver (e.g, forward, backward, etc.) scales cubically as the variables' dimension \cite{MetaOptNet}, it's expensive to derive a solution in the primal space with $O(N^3) \ll O(d^3)$. Moreover, from Equation \ref{eq:rr-dual-itr-weight}, we see that linear combination of feature vectors in the training set drastically alleviate the problem of overfitting. To obtain the final weight, we project $\boldsymbol{a}$ to primal domain by simply applying Equation \ref{eq:rr-dual-itr-weight}.

\noindent\textbf{RR-dual-cls} \cite{R2D2} develops a closed-form solution of RR in dual space for a more discriminative classifier considering the stepwise optimization negates the chances of the model reaching its optimal. The closed-form solution of RR has been successfully applied in tracking in Fourier domain \cite{MOSSE,KCF}, or primal domain \cite{anyobject}, while that in the dual domain is straightforward in our framework by setting the gradient of $L_{meta}$ in Equation \ref{eq:rr-dual-itr} with respect to $\boldsymbol{a}$ to zero, yielding
\begin{equation}
\label{eq:rr-dual-cls}
    \Lambda(\phi(Z); \theta) = \Phi^T(\Phi\Phi^T + \lambda)^{-1}\boldsymbol{y}
\end{equation}
Compared the closed-form solution of Equation \ref{eq:rr-prim-itr}, where $\theta = (\Phi^T\Phi + \lambda)^{-1}\Phi^T\boldsymbol{y}$, matrix $\Phi\Phi^T$ grows quadratically with sample amount $N$ instead of feature dimension $d$ thus drastically alleviate the computation complexity. 

\noindent\textbf{SVM-dual-itr} \cite{MetaOptNet} considers another linear convex classifiers - sparse kernel machine solved in dual space iteratively. While the popularly-applied RR is prone to overfitting and not robust to noisy samples, it only selects a subset of training samples to construct the class boundary. With slack variables $\boldsymbol{\xi}$ introduced, SVM replaces the least-square of residuals in Equation \ref{eq:rr-prim-itr} with a monotonic upper bound of misclassification error, formulated as 
\begin{equation}
\label{eq:svm-dual-itr-misclass}
    \xi_n = \max_k \{w_k^T\phi(x_n) + 1 - \delta_{y_n, k}\} - w_{\widehat y_n}^T\phi(x_n), \forall n
\end{equation}
where $\delta_{\cdot,\cdot}$ is the Kronecker delta function, and $\hat y$ is a real number indexing its class. With weighting term for each sample considered, the few-shot learner is obtained as 
\begin{equation}
\begin{aligned}
\label{eq:svm-dual-itr-prim}
    \Lambda(\phi(Z); \theta) &= \mathop{\arg\min}_{\theta} \sum_{n=1}^N \omega_n \xi_n + \lambda ||\theta||^2 \\
    \mathrm{subject \ to:} \ &\xi_n \ge w_k^T\phi(x_n) + 1 - \delta_{\widehat y_n, k} - w_{\widehat y_n}^T\phi(x_n), \forall n,k.
\end{aligned}
\end{equation}
By adding Lagrange multipliers $\boldsymbol{\eta}$ for inequality constraint and getting the Lagrangian of the above objective, we optimize over the dual variables $\boldsymbol{a}$ and obtain the few-shot learner as 
\begin{equation}
\begin{aligned}
\label{eq:svm-dual-itr}
    \Lambda(\phi(Z); \boldsymbol{\alpha}) = &\mathop{\arg\min}_{\boldsymbol{a}} \boldsymbol{a}^T(\Phi\Phi^T)\boldsymbol{a} - \boldsymbol{a}^T\boldsymbol{y}, \\
    & \mathrm{subject \ to:} \ \alpha_n \le \omega_n y_n \ \mathrm{and} \ \sum_k \alpha_n^k = 0, \forall n,
\end{aligned}
\end{equation}
The complete derivation can be found in \cite{multisvm,MetaOptNet}.

\subsection{Online Tracking}
\label{sec:tracking}

In this section, we illustrate several technical details of our cascaded framework during online tracking. The overall online tracking algorithm is showcased in Algorithm \ref{alg:tracking}.

\noindent\textbf{Candidate Selection.} The candidates sent into the few-shot learner are selected using non-maximal suppression (NMS) with a threshold $0.2$ on the classification and localization result of the first stage without any penalization. Each candidate was applied with RoI Align \cite{maskrcnn} to the size of $5\times5$. For initialization, we draw $24$ samples with groundtruth box appended. Boxes after NMS with the highest IoU with groundtruth are chosen as positives samples, and the rest are negative samples. We further conduct data augmentation (e,g, blurring, rotation, shifting, flip) yielding extra $8$ frames as positive. These samples are used for the initialization of the few-shot learner. During tracking, we draw $M=8$ boxes after NMS per frame and sent them to the few-shot learner. The top-$k$ candidates of the highest fused score $\hat S$ replace the oldest samples in the support set with the highest one being positive while the rest of candidates being negative. By default, $k$ is set to $4$.

\noindent\textbf{Support Set Maintenance.} During online tracking, we maintain a first-in-first-out (FIFO) queue to store the historical frames, with the newest frame replacing the oldest if the memory is full. Along with the features of the candidate $\phi(x_n)$ and its corresponding label $y_n$, we stored a $\omega_n$ to re-weight the samples in optimizing the few-shot learner. Following \cite{ATOM}, we set $w_n$ exponentially decayed with its frame interval to the current frame. The decay rate is set to $0.01$ by default and $0.02$ with distractors detected. For the problem in the primal space, we set the size of memory to $1000$, while for the problem in the dual space, we find out much smaller size around $60$ induces comparable result with acceptable computational cost since the optimization of GPU-based QP solve \cite{OptNet} is of a high demand of CPU resources. To strike a balance between speed and accuracy, we set the size in dual space as $60$. We do not discard the initial augmented positive samples with a minimal weight being $0.15$ since these samples are reliable.

\noindent\textbf{Few-Shot Learner Update.} The few-shot learner update in the primal domain arises naturally with Equation \ref{eq:rr-prim-itr-update}, while for the update in the dual domain, we first cast the dual variables with the current training samples to the primal variables and apply a moving average methodology, formulated as
\begin{equation}
\label{eq:update}
    \theta^{(i+1)} = (1 - \mu) \theta^{(i)} + \mu \Phi_{(i+1)}^T \Lambda(\phi(Z^{(i+1)}); \boldsymbol{a})
\end{equation}
where $\mu$ is a decay rate set as $0.5$ in our experiment by default. $\Phi_{(i+1)}$ and $\phi(Z^{(i+1)})$ denote the design matrix and support set at $i+1$ frame. We update the primal variables $\theta^c$ $10$ recursions in the first frame, and $3$ recursions every $10$ frames or once a distractor is detected, while dual variables set to $10$ recursions and $1$ recursion respectively. 

\setlength{\textfloatsep}{5pt}
\begin{algorithm}[t]
\label{alg:tracking}
\LinesNumbered
\SetKwInOut{Input}{\textbf{Input}}
\SetKwInOut{Output}{\textbf{Output}}
\KwIn{Video sequences $\boldsymbol{f}=\{f^1$,...,$f^L\}$ and initial ground-truth.}
\KwOut{Predicted bounding boxes $\boldsymbol{B}$.}
\BlankLine

Initialize the support set drawn from $f^1$ and augmented positive samples;

Optimize few-shot learner using Equation \ref{eq:rr-prim-itr-update} or \ref{eq:update}; 

\For {i $=$ 2 to $L$}
{
Draw $M$ samples with the first stage and have the second stage solve them;

Check tracking state $\pi$ based on $M$ candidates with fused result $\hat S$ and $\hat B$;

$\boldsymbol{B} \leftarrow \hat B_i$ with the highest $\hat S$;

\If{$\pi\neq$ uncertain \textbf{and} $\pi\neq$ not found}{
Update the support set with top-$k$ samples;
}

\If{$t$ mod $u=0$ \textbf{or} $\pi=$ distractors detected}{
Optimize few-shot learner using Equation \ref{eq:rr-prim-itr-update} or \ref{eq:update}; 
}
}
\caption{Online tracking Algorithm.}
\end{algorithm}

\noindent\textbf{Stage Fusion.} For each type of the few-shot learner in the second stage, we have a confidence score $S_{meta}$ for the candidates proposed in the first stage, with a refined localization using boxes given in the first stage as anchors. Since the architecture of the localization branch consists of FC layers following traditional R-CNN \cite{fasterrcnn}, we denote it as $B_{rcnn}$. Let $S_{rpn}$ and $B_{rpn}$ be the classification and localization result of these candidates in the first stage. Here $S_{rpn}$ is penalized with size change, ratio change, and cosine window \cite{SiamRPN}. Following \cite{SPM}, the fused score $\hat S$ used to sort the candidates and their corresponding bounding boxes $\hat B$ can be derived as
\begin{equation}
\begin{aligned}
\label{eq:fusion}
    \hat S &= (1 - \mu_{cls}) S_{rpn} + \mu_{cls}S_{meta} \\
    \hat B &= \frac{S_{rpn}}{\mu_{loc}S_{meta} + S_{rpn}}B_{rpn} + \frac{\mu_{loc}S_{meta}}{\mu_{loc}S_{meta} + S_{rpn}}B_{rcnn}
\end{aligned}
\end{equation}
where $\mu_{cls}$ and $\mu_{loc}$ are the fusion weights and is set with heuristics or hyper-parameter grid search.

\section{Implementation Details}

\noindent\textbf{Training.} Similar to \cite{DiMP}, to achieve the end-to-end training, we sample data in a video manner with a training (support) set $Z$ and a testing (query) set $Z'$, each with $3$ and $2$ frames from a video clip with interval less than $100$. The backbone is modified to original ResNet-50 \cite{ResNet} with the features from the last $3$ layers interpolated to the same size. For the second stage, we draw $16$ samples based on the result of the first stage after NMS with the threshold being $0.1$ by default. Groundtruth boxes are appended to ensure training stability. Boxes with IoU with groundtruth boxes higher than $0.8$ are assigned as positive samples and negative if lower than $0.2$. Total $8$ samples with maximal $2$ positive samples are sampled per image. Note that the label assignment on both meta-training and meta-testing follows the same above procedure. The search region is set to $255 \times 255$ by default while the template is set to $127 \times 127$. We modify the first-stage SiamRPN++ using ATSS \cite{atss} for label assignment, which slightly improve the performance of SiamRPN++ (see Table \ref{tab:ablation} for details).

We use SGD with the learning rate exponentially decayed from $5e-3$ to $5e-4$ in $20$ epochs, which taskes $12$ hours with $16$ GTX 1080ti GPUs. Warm-up for the first $5$ stage is used with the learning rate first grows from $1e-3$ to $5e-3$ at a linear pace. The backbone ResNet-50 is frozen for the first $10$ epochs and updated with a $10$ times smaller learning rate. The training datasets include: ImageNet VID+DET \cite{imagenet}, Youtube-BB \cite{youtubebb}, and COCO \cite{coco}.

\noindent\textbf{Total Loss.} With two stages together, we formulate our offline-training loss function $L$ as follows:
\begin{equation}
\begin{aligned}
L = & \frac{a_3}{N_{sec}^b}\sum_n L_{sec}^{cls}(c'_n,\hat c'_n) + \frac{1}{N_{fir}^b} \sum_{x,y}\{ a_1 L_{fir}^{cls}(c_{x,y}, \hat c_{x,y}) \\
& + \mathbbm{1}_{\{\hat c_{x,y}\textgreater 0\}}[a_2 L_{fir}^{loc}(r_{x,y},\hat r_{x,y}) + a_4 L_{sec}^{loc}(r'_{x,y},\hat r'_{x,y})]\}
\end{aligned}
\end{equation}
where $L_{fir}^{cls}$ and $L_{sec}^{cls}$ are the Focal Loss \cite{RetinaNet} with negative and positive samples split and contributing to final loss equally to avoid class imbalance. $L_{fir}^{loc}$ and $L_{sec}^{loc}$ are the L1 loss. $N_{fir}^b$ denotes the number of positive samples in first stage per batch. $\mathbbm{1}_{\{\hat c_{x,y}\}}$ is the indicator function, being 1 if $\hat c_{i} > 0$ and $0$ otherwise. We train the localization branch in the second stage with positives assigned in the first stage instead of the positives to train the few-shot learner. The re-weighting factors $a_1$, $a_2$, $a_3$, $a_4$ is set to $10$, $5$, and $1.2$, and $0.6$ respectively.

\section{Experiments}
In this section, we first investigate the effectiveness of the few-shot learners and several technical details. Further, we evaluate our proposed method on the major benchmarks and compare the results with previous trackers. Speed for all experiments is reported on an NVIDIA Titan Xp GPU. Detailed results, code, and video demos will be made available upon acceptance. 

\subsection{Ablation Study}

\begin{figure*}[!t]
\centering
\includegraphics[width=17.8cm]{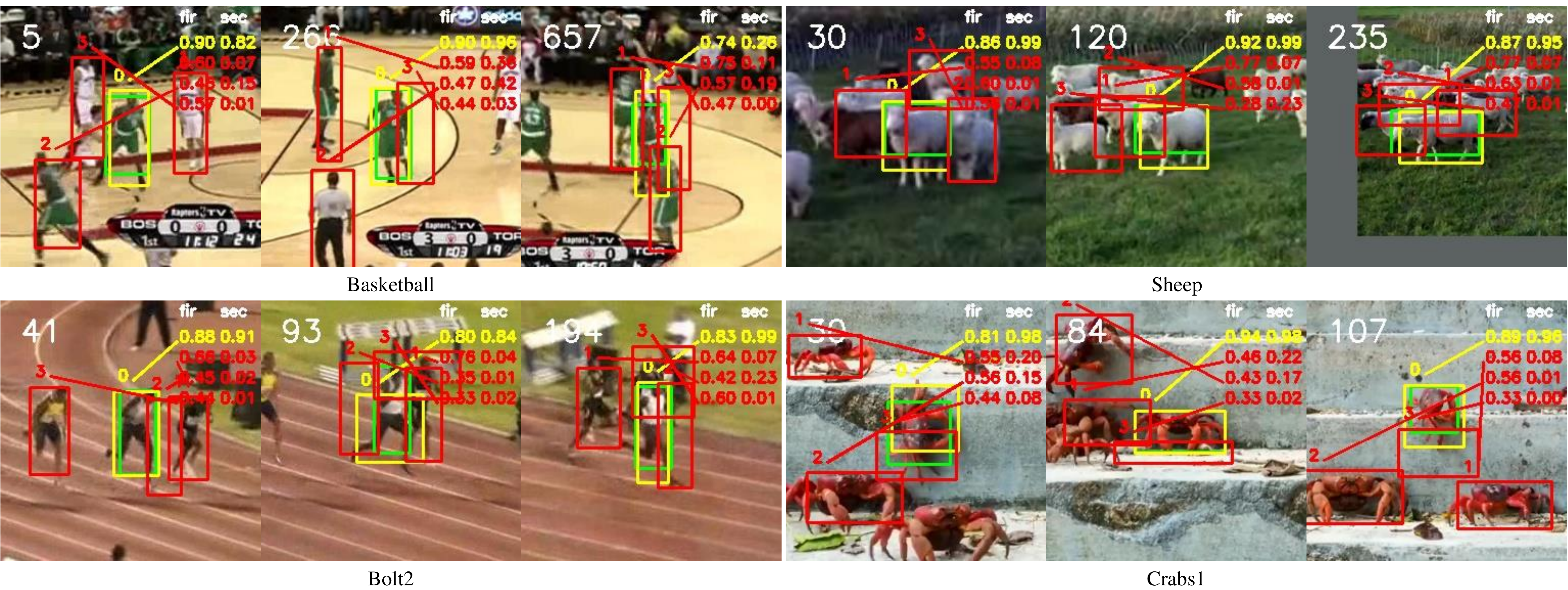}
\caption{Visualization of the confidence score form the first stage and the second stage. Four sequences from VOT2018 benchmark-Basketball, Sheep, Bolt2, and crabs1-are showcased. In the upper right corner of each frame, the confidence scores from two stages are listed, with the first column being the matching score from proposal network and the second column being the classification score from the few-shot learner. The index annotated on each bounding box is sorted according to the final fused score $\hat S$. \textcolor{green}{Green} denotes the groundtruth. \textcolor{yellow}{Yellow} denotes the predicted target. \textcolor{red}{Red} denotes the predicted negative candidates. Best viewed with color and amplification.}
\label{visualization}
\vspace{-0.5cm}
\end{figure*}

We perform extensive ablation studies to showcase the validity of our proposed method. The experiments are conducted on the dataset of VOT2018 and NUO323 (a combination of NFS100 (30fps), UAV123, and OTB100).

\begin{table}[!t]
\setlength{\tabcolsep}{0.7mm}{
\centering
\begin{tabu}{c|c|c|c|c|c|c} 
\multirow{2}{*}{second stage} & \multicolumn{3}{c|}{VOT2018} & \multicolumn{2}{c|}{NUO323} & \multirow{2}{*}{FPS} \\  
\cline{2-6}
& Acc & R & EAO & AUC & NPr \\
\Xhline{1.2pt} 
none \cite{SiamRPN++} & 0.603 & 0.201 & 0.414 & 0.603 & 0.770 & \multirow{2}{*}{\bf{90}} \\
none$^\dag$ \cite{SiamRPN++} & 0.598 & 0.192 & 0.432 & 0.609 & 0.777 & \\
\hline	
fc-dotproduct & 0.583 & 0.258 & 0.383 & 0.597 & 0.759 & 58 \\
fc-euclidean & 0.593 & 0.220 & 0.426 & 0.601 & 0.771 & 57 \\
fc-cosine & 0.594 & 0.225 & 0.418 & 0.600 & 0.768 & 55 \\
fc-relation & 0.601 & 0.215 & 0.420 & 0.599 & 0.759 & 58 \\
\hline
MatchingNet \cite{matching} & 0.586 & 0.196 & 0.445 & 0.619 & 0.793 & 57 \\
ProtoNet \cite{prototypical} & 0.573 & 0.183 & 0.439 & 0.621 & 0.796 & 56 \\
\hline
RR-prim-itr \cite{MAML} & 0.601 & 0.117 & 0.510 & 0.644 & 0.829 & 40 \\
RR-dual-itr \cite{MetaOptNet} & \bf 0.619 & \bf 0.122 & \bf 0.529 & 0.642 & 0.825 & 45 \\
RR-dual-cls \cite{R2D2}& 0.608 & 0.131 & 0.509 & 0.646 & 0.824 & 56 \\
SVM-dual-itr \cite{MetaOptNet} & 0.605 & 0.122 & 0.517 & \bf 0.647 & \bf 0.830 & 50 \\
\end{tabu}}
\vspace{0.05cm}
\caption{Ablation study on the methods of the second stage for classification. \textit{none} denotes no second stage is cascaded. \dag \ denotes our modified version.}
\label{tab:ablation}
\end{table} 

\noindent\textbf{Relations with Other Cascaded Pipelines.} The multi-stage cascaded architecture has been introduced in the field of visual tracking, while our method is the first method to replace the fully convolutional (FC) pipeline with a differentiable optimization submodule. Intuitively, such difference shares a similar relationship with metrics-based methods and optimization-based methods in the field of few-shot classification. To serve as a baseline for comparison, we also implement several FC counterparts with different distance metrics. Specifically, we conduct experiments on $4$ type of metrics: (1) Dot Product, (2) Cosine Distance (3) Euclidean Distance, and (4) Relation Network, dubbed as \textit{fc-dotproduct}, \textit{fc-euclidean}, \textit{fc-cosine}, and \textit{fc-relation} respectively. The detailed results are showcased in Table \ref{tab:ablation}.

\noindent\textbf{Other Genre of Few-Shot Learners.}
Since the second stage \ref{sec:secondstage} in our framework is designed without specification of FSC approaches, thus large variety of FSC algorithm could be incorporated in our framework. Specifically, we consider metric-based methodologies Matching Network \cite{matching} and Prototypical Network \cite{prototypical}, dubbed at \textit{MatchingNet} and \textit{ProtoNet}, which classifies sample based the cosine/euclidean distance between the sample and the class prototypes.
Note that \textit{MacthingNet} and \textit{ProtoNet} differs from \textit{fc-cosine} and \textit{fc-euclidean} in having an online learning pipeline and a memory update mechanism.
The detailed results are showcased in Table \ref{tab:ablation},

\begin{table}[!t]
\setlength{\tabcolsep}{0.7mm}{
\centering
\begin{tabu}{c|c|c|c|c|c} 
\multirow{2}{*}{first stage} & \multirow{2}{*}{second stage} & \multicolumn{3}{c|}{VOT2018} & \multirow{2}{*}{FPS} \\  
\cline{3-5}
& & Acc & R & EAO \\
\Xhline{1.2pt}
\multirow{2}{*}{DiMP \cite{DiMP}} 
 & none$^\dag$ & 0.582 & 0.117 & 0.499 & 42 \\
 & SVM-dual-itr & 0.600 & \bf{0.108} & \bf{0.523} & 28 \\
\hline
\multirow{2}{*}{SiamRPN++ \cite{SiamRPN++}} 
 & none$^\dag$ & 0.598 & 0.192 & 0.432 & \bf{90} \\
 & SVM-dual-itr & \bf{0.605} & 0.117 & 0.517 & 50 \\
\end{tabu}}
\vspace{0.05cm}
\caption{Ablation study on the method of the first stage for candidate proposal. \textit{none} denotes no second stage is cascaded. \dag \ denotes our modified version for fair comparison.}
\label{tab:ablation2}
\end{table} 

\noindent\textbf{Effectiveness of Siamese Proposals.} We note that reference \cite{CARE} also develops a cascaded tracking framework with the first stage as recently the proposed ATOM \cite{ATOM}. In contrast, its follow-up work DiMP \cite{DiMP} which entails the end-to-end training in an optimization-based few-shot learning fashion is evaluated in our framework. For a fair comparison, we discard the regression method proposed in \cite{ATOM, DiMP} while resorting to \cite{SiamRPN++} in the same setting as above. The mere difference, therefore, lies in the heatmap generation approach. For succinctness, we only showcase the result of \textit{SVM-dual-itr} in the following ablation study without specification. From Table \ref{tab:ablation2}, we find that a DiMP-alike first stage robustly yields high-quality candidates as well but operates at a speed $3$ times slower than a siamese region proposal network.
Moreover, when no second stage is cascaded, the online classifier as the first stage similar to DiMP ostentatiously exceeds SiamRPN-alike template matching by a large margin, while the gap is greatly eliminated by introducing a second-stage few-shot learner.

\begin{table}[!t]
\setlength{\tabcolsep}{1.6mm}{
\centering
\begin{tabu}{c|c|c|c|c|c|c} 
\multirow{2}{*}{memory} & \multicolumn{3}{c|}{VOT2018} & \multicolumn{2}{c|}{NUO323} & \multirow{2}{*}{FPS} \\  
\cline{2-6}
& Acc & R & EAO & AUC & NPr \\
\Xhline{1.2pt} 
$M=200$ & \bf{0.615} & 0.164 & 0.451 & 0.643 & 0.826 & 38 \\
$M=100$ & 0.590 & 0.136 & 0.469 & \bf{0.647} & \bf{0.830} & 45 \\
$M=80$ & 0.592 & 0.155 & 0.480 & 0.639 & 0.819 & 48 \\
$M=60$ & 0.605 & \bf{0.122} & \bf{0.517} & 0.633 & 0.813 & 50 \\
$M=40$ & 0.611 & 0.159 & 0.478 & 0.634 & 0.812 & \bf{52} \\
\end{tabu}}
\vspace{0.05cm}
\caption{Ablation study on the memory size $M$. The result of \textit{SVM-dual-itr} is reported for briefness, with other few-shot learner following a similar trend.}
\label{tab:ablation3}
\end{table} 

\noindent\textbf{Memory of Support Set.} To strike a balance between speed and accuracy, we investigate the influence memory size exerts on the tracker performance. By default, the memory of $60$ candidates is applied (i.e, for VOT2018 dataset) while a memory size of $100$ reaches optimal results for the dataset of NUO323, presumably due to large deformation of VOT2018 precluding the benefits of long-term memory. Moreover, the robustness of the tracker is still warranted when memory size dwindles to a minimal capacity of $40$ candidates. Such trade-off can be salutary to real-world applications. 

\noindent\textbf{Visualization of Candidate Confidence.} To validate the effectiveness of the proposed method, we offer a quantitative analysis of the confidence score of the candidates in several video sequences, shown as Figure \ref{visualization}. Comparing two columns from given frames, we find that the confidence scores from the second stage are more concentrated toward the target thus more robust to the distractors. In some cases (i.e, the third frame from the sequence Basketball), when the siamese network fails to distinguish between the target and the distractors, the few-shot learner can effectively avoid the tracker from drifting. Therefore, the effectiveness of the few-shot learner for more discriminative power can be more clearly observed.

\subsection{Comparison with State-of-the-Arts}
We compare our proposed method, termed \textbf{FsTrack}, with state-of-the-art approaches on the major tracking benchmarks. The result of \textit{SVM-dual-itr} is reported for briefness.

\begin{table}[!t]
\footnotesize
\setlength{\tabcolsep}{0.3mm}{
\centering
\begin{tabu}{c|c|c|c|c|c|c|c|c|c} 
 & \tabincell{c}{UPDT \\ \cite{UPDT}} & \tabincell{c}{SiamRPN \\ ++\cite{SiamRPN++}} & \tabincell{c}{ATOM \\ \cite{ATOM}} & \tabincell{c}{DiMP \\ \cite{DiMP}}  & \tabincell{c}{DROL \\ \cite{DROL}} & \tabincell{c}{Ocean \\ \cite{Ocean}} & \tabincell{c}{FCOT \\ \cite{FCOT}} & \tabincell{c}{RPT \\ \cite{RPT}} & Ours \\
\Xhline{0.8pt}
A$\uparrow$ & 0.536 & 0.600 & 0.590 & 0.597 & \green{0.616} & 0.592 & 0.600 & \red{0.629} & \blue{0.605} \\
R$\downarrow$ & 0.184 & 0.234 & 0.204 & 0.153 & 0.122 & \blue{0.117} & \green{0.108} & \red{0.103} & \blue{0.117} \\
EAO$\uparrow$ & 0.378 & 0.414 & 0.401 & 0.440 & 0.481 & 0.489 & \blue{0.508} & \green{0.510} & \red{0.517} \\
\end{tabu}}
\vspace{0.05cm}
\caption{Results on VOT2018 challenge dataset \cite{vot2018} in terms of expected average overlap (EAO), robustness (R), and accuracy (A). \red{Red}, \green{green} and \blue{blue} denote top-$3$ results.} 
\label{tab:vot2018}
\end{table} 

\noindent\textbf{VOT2018 \cite{vot2018}.} VOT2018 dataset consists of $60$ challenging videos. Trackers are restarted at failure according to its protocol. The tracker's overall performance is evaluated upon robustness and accuracy, defined using failure rate and IoU and a comprehensive protocol EAO involves both two respects. We compare our methods with the state-of-the-art methods as shown in Table \ref{tab:vot2018}, resulting a huge performance gain on EAO from $0.510$ to $0.0.517$ with a $1.9\%$ relative gain. Moreover, our tracker achieve a comparable robustness and accuracy with previous methods.

\begin{table}[!t]
\footnotesize
\setlength{\tabcolsep}{0.05mm}{
\centering
\begin{tabu}{c|c|c|c|c|c|c|c|c|c} 
 & \tabincell{c}{MDNet \\ \cite{MDNet}} & \tabincell{c}{ECO \\ \cite{ECO}} & \tabincell{c}{SiamRPN \\ ++\cite{SiamRPN++}} & \tabincell{c}{ATOM \\ \cite{ATOM}}  & \tabincell{c}{Ocean \\ \cite{Ocean}} & \tabincell{c}{DiMP \\ \cite{DiMP}} & \tabincell{c}{FCOT \\ \cite{FCOT}} & \tabincell{c}{UPDT \\ \cite{UPDT}} & Ours \\
\Xhline{0.8pt}
OTB2015 & 67.8 & 69.1 & \blue{69.6} & 66.9 & 68.4 & 68.4 & 69.3 & \red{70.2} & \green{69.7} \\
NFS (30fps) & 42.2 & 46.6 & - & \blue{58.4} & - & \green{62.0} & - & 53.7 & \red{62.2} \\
\end{tabu}}
\vspace{0.05cm}
\caption{Results on OTB2015 and NFS (30fps) datasets \cite{otb2015,nfs} in terms of overall AUC score. } 
\label{tab:otbnfs}
\end{table} 

\noindent\textbf{OTB2015 \cite{otb2015}.} OTB2015 dataset contains a total amount of $100$ sequences with motion, scale change, and illumination change, etc. The evaluation protocol is over precision plot and success plot (AUC). As shown in Table \ref{tab:otbnfs}, we achieve a comparable performance with an AUC of $69.7$ and a precision of $92.2$ compared with the top-leading trackers. By achieving a top result in terms of success rate, the validity of our proposed method is better demonstrated.

\noindent\textbf{NFS \cite{nfs}.} The $30$ fps version of dataset Need for Speed (NFS) resembles the evaluation protocol of OTB2015. With fast motion of targets, along with challenging scenarios like distractors and scale change, NFS functions well for a comprehensive evaluation benchmark. As shown in Table \ref{tab:otbnfs}, our tracker achieves an AUC score of $62.2$ and a precision score of $74.7$, indicating an desirable tracking capability.


\begin{figure}[!t]
\centering
\subfigure{\includegraphics[width=0.49\linewidth,height=0.42\linewidth]{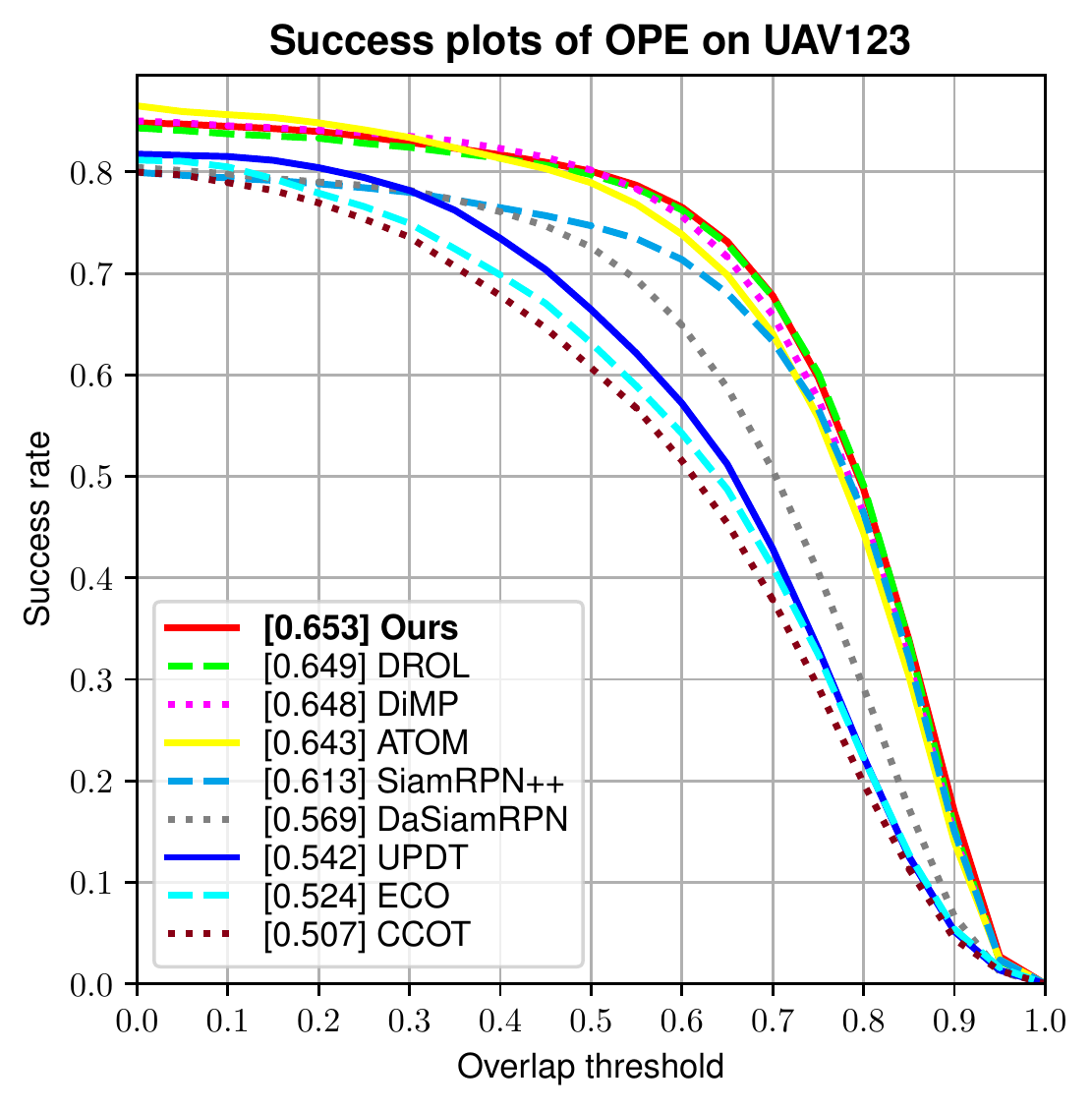}}
\subfigure{\includegraphics[width=0.49\linewidth,height=0.42\linewidth]{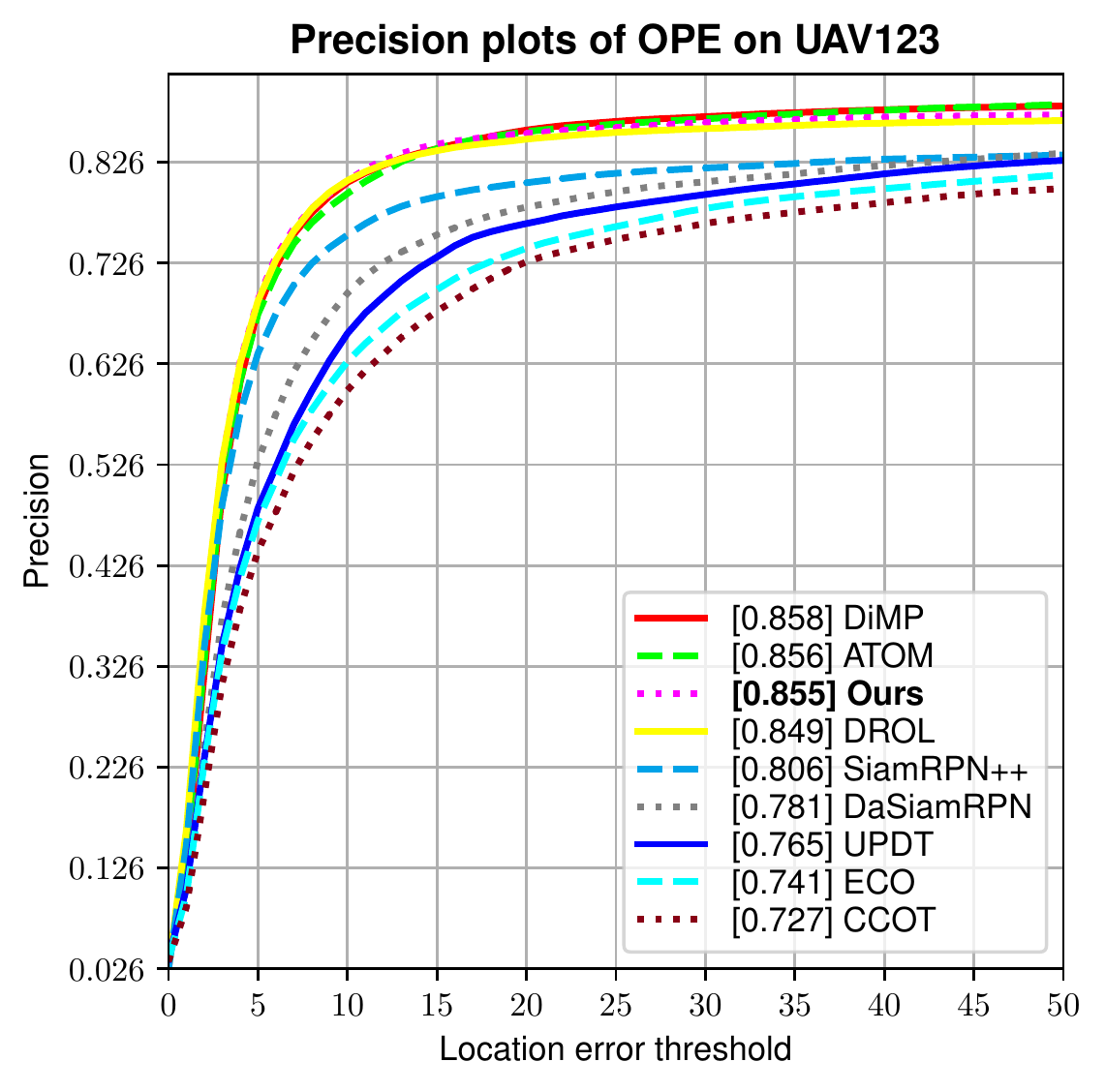}}
\vspace{-0.05cm}
\caption{Results on UAV123 dataset \cite{uav123} in terms of overall AUC score (left) and precision score (right).}
\label{fig:uav}
\end{figure}

\noindent\textbf{UAV123 \cite{uav123}.} UAV123 dataset contains $123$ sequences collected from a UAV perspective, which are practical in reality considering one major application of tracking algorithm is for UAVs' purpose. As shown in \ref{fig:uav}, our tracker achieved an AUC score of $65.3$ and a precision score of $85.5$, with a large performance gain in overall success rate and a comprable performance in terms of precision score.


\begin{table}[!t]
\footnotesize
\setlength{\tabcolsep}{0.1mm}{
\centering
\begin{tabu}{c|c|c|c|c|c|c|c|c|c} 
 & \tabincell{c}{SiamFC \\ \cite{SiamFC}}& \tabincell{c}{UPDT \\ \cite{UPDT}} & \tabincell{c}{UpdateN \\ et\cite{UpdateNet}} & \tabincell{c}{ATOM \\ \cite{ATOM}} & \tabincell{c}{SPM \\ \cite{SPM}}  & \tabincell{c}{SiamRPN \\ ++\cite{SiamRPN++}} & \tabincell{c}{DiMP \\ \cite{DiMP}} & \tabincell{c}{DCFST \\ \cite{DCFST}} & Ours \\
\Xhline{0.8pt}
AUC$\uparrow$ & 57.1 & 61.1 & 67.7 & 70.3 & 71.2 & 73.3 & \green{74.0} & \red{75.2} & \green{73.6} \\
NPr$\uparrow$  & 66.6 & 70.2 & 75.2 & 77.1 & 77.8 & 80.0 & \blue{80.1} & \green{80.9} & \red{81.3} \\
Pr$\uparrow$ & 53.3 & 55.7 & 62.5 & 64.8 & 66.8 & \blue{69.4} & 68.7 & \green{70.0} & \red{70.2} \\
\end{tabu}}
\vspace{0.05cm}
\caption{Results on TrackingNet test set \cite{trackingnet} in terms of precision (Pr), normalized precision (NPr), and success (AUC). } 
\label{tab:trackingnet}
\end{table} 

\noindent\textbf{TrackingNet \cite{trackingnet}.} TrackingNet is a large-scale tracking benchmark, which consists over $30k$ sequences in total and $511$ sequences for testing without publicly available groundtruth. The results, shown in \ref{tab:trackingnet}, demonstrates that our proposed method achieve a comparable performance in AUC, normalized precision, and precision, validating the effectiveness of our proposed method.

\begin{table}[!t]
\footnotesize
\setlength{\tabcolsep}{0.1mm}{
\centering
\begin{tabu}{c|c|c|c|c|c|c|c|c|c} 
 & \tabincell{c}{SiamFC \\ \cite{SiamFC}} & \tabincell{c}{SiamRPN \\ ++\cite{SiamRPN++}} & \tabincell{c}{SPM \\ \cite{SPM}} & \tabincell{c}{ATOM \\ \cite{ATOM}} & \tabincell{c}{DMV \\ \cite{DMV}} & \tabincell{c}{DCFST \\ \cite{DCFST}} & \tabincell{c}{DiMP \\ \cite{DiMP}} & \tabincell{c}{Ocean \\ \cite{Ocean}} & Ours \\
\Xhline{0.8pt}
$\mathrm{SR}_{0.50}\uparrow$ & 35.3 & 61.8 & 59.3 & 63.4 & 69.5 & 71.6 & \blue{71.7} & \red{72.1} & \green{72.0} \\
$\mathrm{SR}_{0.75}\uparrow$ & 9.8 & 32.9 & 35.9 & 40.2 & 49.2 & 46.3 & \red{49.2} & \blue{48.7} & \green{49.1} \\
AO$\uparrow$ & 34.8 & 51.8 & 51.3 & 55.6 & 60.1 & \blue{61.0} & \green{61.1} & \green{61.1}  & \red{61.2} \\
\end{tabu}}
\vspace{0.05cm}
\caption{Results on GOT-10k test set \cite{got10k} in terms of average overlap (AO), $\mathrm{SR}_{0.75}$, and $\mathrm{SR}_{0.5}$.} 
\label{tab:got10k}
\end{table} 

\noindent\textbf{GOT-10k \cite{got10k}.} GOT-10k is a recently-proposed large-scale dataset for both training and testing, with no overlap in object classes between training and testing. For GOT-10k test, we train our tracker by only using the GOT10k train split following its standard protocol. From Table \ref{tab:got10k}, our proposed method achieve comparable results with the top methods with an AO of $61.2$ and a $\mathrm{SR}_{0.50}$ of $72.0$.

\section{Conclusion}

In this work, we propose a generalized two-stage cascaded framework to enable versatile few-shot learning for real-time visual tracking. 
The first stage filters out easy negatives while retaining a high recall on all possible candidates via a siamese matching network, and the second stage distinguishes among these potential candidates by solving a few-shot classification problem. 
In demand of a real-time speed and fast adaptability towards novel classes, this work mainly focuses on the discussion of optimization-based few-shot learners. However, the proposed coarse-to-fine cascaded framework does not limit to any specific genre of FSL algorithms, since proposed sparse samples suit better for the task-setting of FSL, which will strongly strength the connections between visual tracking and few-shot learning communities.
The effectiveness of our method is validated among the major benchmarks with a large margin induced and a speed beyond real-time requirement. 

{\small
\bibliographystyle{ieee_fullname}
\bibliography{egbib}
}

\end{document}